\documentclass[letterpaper, 10 pt, conference]{ieeeconf}
\IEEEoverridecommandlockouts
\usepackage{cite}
\usepackage{amsmath,amssymb,amsfonts}
\usepackage{graphicx}
\usepackage{textcomp}
\usepackage{comment}
\def\BibTeX{{\rm B\kern-.05em{\sc i\kern-.025em b}\kern-.08em
    T\kern-.1667em\lower.7ex\hbox{E}\kern-.125emX}}
    
\usepackage{algorithm}
\usepackage{algpseudocode}
\makeatletter
\def\BState{\State\hskip-\ALG@thistlm}
\makeatother
\usepackage[bottom]{footmisc}
\usepackage{xcolor}
\usepackage{cellspace} 
\usepackage{makecell} 
\setcellgapes{2pt}
\usepackage{booktabs}

\begin{document}

%title
\title{Optimal Dexterity for a Snake-like Surgical Manipulator using Patient-specific Task-space Constraints in a Computational Design Algorithm\\
}

%authors
%\author{\IEEEauthorblockN{Andrew Razjigaev}
%, Liao Wu, Jonathan Roberts, Ross Crawford,  Ajay K. Pandey

% \IEEEauthorblockA{\textit{Australian Centre for Robotic Vision} \\
% \textit{Science and Engineering Faculty}\\
% \textit{Queensland University of Technology}\\
% Brisbane, Australia \\
% Email: a.razjigaev@qut.edu.au}

% \thanks{This work was supported in part by the Australian Government Research Training Program (RTP) Stipend (Domestic) and the Australian Centre for Robotic Vision (ACRV) Top-Up Scholarship.  Computational resources and services used in this work were provided by the HPC and Research Support Group, Queensland University of Technology, Brisbane, Australia}

% }
%authors
\author{Andrew~Razjigaev, Ajay K. Pandey, Jonathan Roberts~\IEEEmembership{Senior Member~IEEE}, and Liao~Wu~\IEEEmembership{Member~IEEE}% <-this % stops a space
	%	\thanks{This work was supported in part by the Australian Government Research Training Program (RTP) Stipend (Domestic) and the Australian Centre for Robotic Vision Top-Up Scholarship.  Computational resources and services used in this work were provided by the HPC and Research Support Group, Queensland University of Technology, Brisbane, Australia}
    %\thanks{A. Razjigaev is with the Australian Centre for Robotic Vision, Science and Engineering Faculty, Queensland University of Technology, Brisbane, Australia. {\tt\small a.razjigaev@qut.edu.au}}
	\thanks{A. Razjigaev, A. K. Pandey, and J. Roberts are with the Science and Engineering Faculty at Queensland University of Technology and the Australian Centre for Robotic Vision, Brisbane, Australia. {\tt\small a.razjigaev@qut.edu.au}}
	\thanks{L. Wu is with the School of Mechanical and Manufacturing Engineering, University of New South Wales, Sydney, Australia. {\tt\small dr.liao.wu@ieee.org}}
}

\maketitle

%document paragraphs
\begin{abstract}
Tendon-driven snake-like arms have been used to create highly dexterous continuum robots so that they can bend around anatomical obstacles to access clinical targets. In this paper, we propose a design algorithm for developing patient-specific surgical continuum manipulators optimized for oriental dexterity constrained by task-space obstacles. The algorithm uses a sampling-based approach to finding the dexterity distribution in the workspace discretized by voxels. The oriental dexterity measured in the region of interest in the task-space formed a fitness function to be optimized through differential evolution. This was implemented in the design of a tendon-driven manipulator for knee arthroscopy. The results showed a feasible design that achieves significantly better dexterity than a rigid tool. This highlights the potential of the proposed method to be used in the process of designing dexterous surgical manipulators in the field.
\end{abstract}

% \begin{IEEEkeywords}
% tendon Driven Manipulator, Continuum robot, Optimal Design, Patient-specific, Evolution algorithm, Snake-like robot, variable neutral line mechanism, Computational design, numerical algorithms
% \end{IEEEkeywords}

\section{Introduction}
%\textcolor{red}{I found a few grammatical errors. Suggest using tools like Grammarly to check. - Leo}

Tendons have been used to drive a variety of snake-like continuum and pseudo-continuum (hyper-redundant) robotic arms in the field. Minimally invasive surgery is one application where these robots are greatly beneficial because they are able to reach difficult-to-access surgical sites and complete tasks with great dexterity. Tendon-driven snake-like robots allow actuation to be done extrinsically from the arm allowing these robots to be miniaturized. These robots can decrease the footprint of teleoperated robotic surgery significantly as opposed to the current straight and rigid tools of state-of-the-art systems like the da Vinci \cite{Burgner-Kahrs2015}.

\begin{figure}[htbp]
	\centerline{\includegraphics[width=85mm]{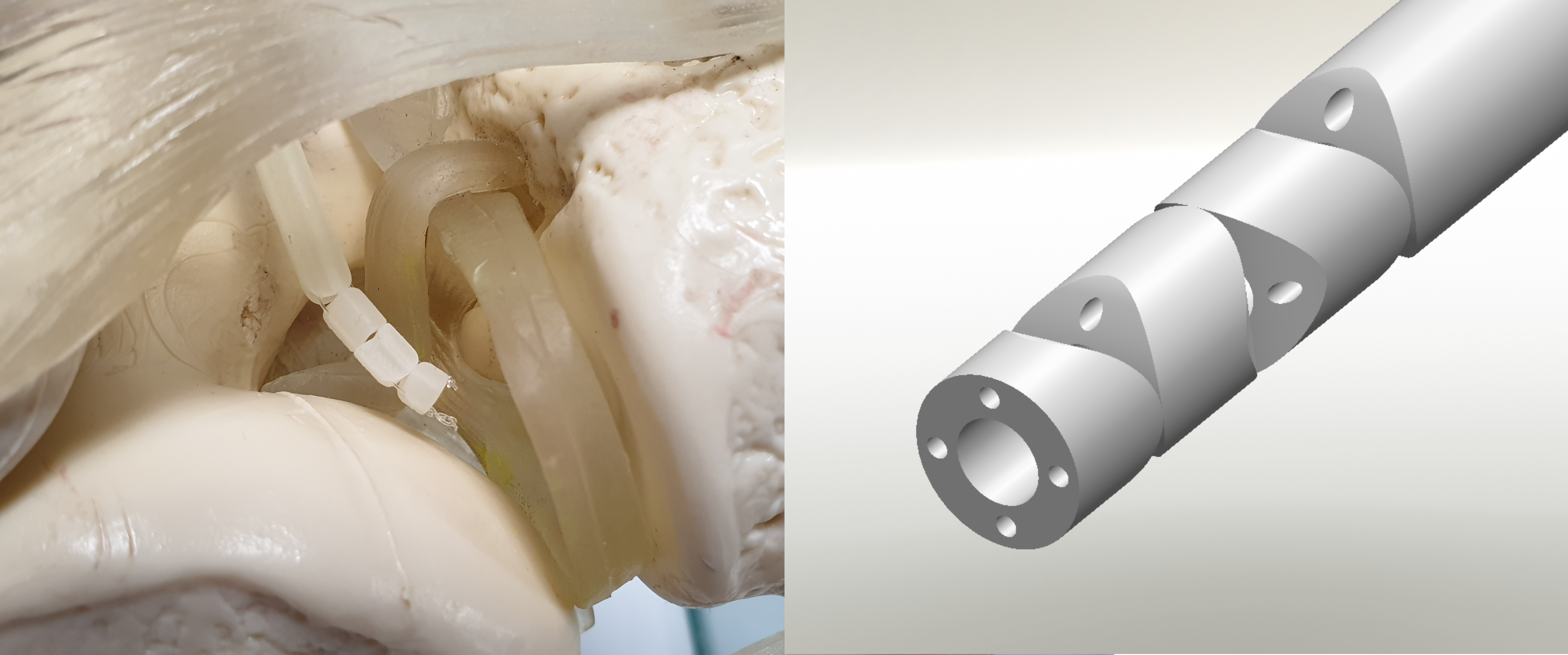}}
	\caption{The optimal design of a dexterous tendon driven snake-like robot for knee arthroscopy as a CAD model (right) and as a 3D print inside a model knee (left)}
	\label{fig1}
\end{figure}

In teleoperated surgery, dexterity is particularly important because the robot is not only required to reach the surgical sites but is also required to perform complicated manipulation. A recent survey on orthopaedic surgeries \cite{Anjali2017} revealed that surgeons find the procedure ergonomically challenging and have frustration with the current tools and technology. Dexterous snake-like robots controlled from a teleoperation console can be an ergonomic solution to the problems in knee arthroscopy. Therefore, this paper proposes a computational design algorithm for developing patient-specific snake-like tools for dexterous manipulation in arthroscopy, as illustrated in Fig. \ref{fig1}. 
%This paper also implements the algorithm in the surgical case study of designing a pseudo-continuum manipulator for knee arthroscopic teleoperation.

\subsection{Related Work}

Optimizing the design for snake-like robotic arms has often involved finding trade-offs in relationships with the mechanical design and its task space performance. The initial studies for optimal snake-like robots involved trade-offs with the workspace and certain mechanical properties like the stiffness in a flexible backbone joint \cite{Zhang2010}. When the patient-specific paradigm emerged, many literature in the field developed cost-functions to describe trade-offs for designing continuum robots for better workspaces and path-planning in patient-specific anatomical environments. This was strongly evident in the design of concentric tube robots where patient 3D anatomies were collected by Magnetic Resonance Images (MRI) and optimizations were done to find the minimal tube lengths and their curvatures while penalizing collisions in a simulated navigation task to some desired targets \cite{Bedell2011} \cite{Anor2011}. 
In other literature, a sampling-based motion planning approach to the problem was proposed to avoid complex inverse kinematics calculations \cite{Torres2012}. 

Later in the field, new algorithms appeared that focused more towards on task space reachability in a volume rather than navigation to a point. Among those works were methods that discretised the workspace into voxels and generated an objective function to maximize the coverage of the concentric tube workspace in the region of interest \cite{Burgner2013}. 
Other algorithms turned this approach into an occupancy grid map where obstacles were dilated and sampled configurations of a centre-line representation of the robot were used to develop a cost function for motion planning \cite{Bergeles2015}. One algorithm maximized reachability using a sampling-based motion planner to enable motions to a variety of target points in the lung \cite{Baykal2015}. All of these methods for optimizing continuum robots involve defining a cost function that would be optimized using generic optimization algorithms like generalized pattern search, the Nelder-Mead simplex algorithm and adaptive simulated annealing. These studies provided continuum robot algorithms for better navigation and reachability in patient-specific anatomies but none of them focused on optimizing for dexterity.

Other patient-specific robot studies focused on the development of cost-effective disposable tools. 
Some of these works include developing patient-specific 3D printable concentric tube robots \cite{Amanov2015} from 3D Ultrasound scans \cite{Morimoto2017}. One study designed disposable tendon-driven end-effectors for concentric tube robots to increase the dexterity at the tip \cite{Prasai2016}. In addition to this study, there was further work that conducted a dexterity analysis for the combination of disposable tendon-driven end-effectors with concentric tube robots \cite{Liao2017}. They found that adding more degrees of freedom (DOF) to the distal end of the robot increased dexterity while adding DOF to the proximal end increased the workspace. They used the concept of orientability for analyzing dexterity which could be used as a guideline for future continuum robot design.

Other fields of research with conventional robotic manipulators had used the measure for manipulability for dexterity which is the singular value decomposition of the Jacobian matrix for a manipulator \cite{Jin1991} \cite{Abdel2004}. One study with a conventional manipulator used a genetic algorithm to minimize a cost function to find the best trade-off between manipulability and the distance from obstacles in path-planning \cite{Lehnert2015}. Although manipulability provides a quick and easy calculation for dexterity, the measure is only dependent on the joints and the forward kinematics which may not be suitable for dexterity in confined environments with many obstacles.
Overall, the literature demonstrates that continuum robots are becoming patient-specific, disposable and optimized navigation and reachability performance in anatomical structures. However, there is a lack of work in designing them for the task performance of oriental dexterity in anatomical structures.

\subsection{Contribution}
Although there are computational design algorithms that exist for snake-like robots, no literature - as far as the authors are aware - has used oriental dexterity as a metric for their design given a confined task-space environment. Therefore, the contribution of this paper is to propose a novel computational algorithm for the optimal design of a robot arm for dexterous teleoperation. This algorithm was implemented for the surgical case study of designing a pseudo-continuum manipulator for knee arthroscopic teleoperation.

\section{Design Problem Formulation}
The problem is to create a process for designing dexterous robotic manipulators as a function of the task space. Specifically, it is to find the parameters for the design of a snake-like robot such that the dexterity is highest in the region of interest taken account of task space obstacles. 
The region of interest is the objects in the task space that are meant to be manipulated by the robot arm represented as voxels.

Given a set of parameters within bounds, all that can be provided from the robot arm model are:
\begin{itemize}
  \item Samples of the configuration space bounded by joint limits;
  \item The forward kinematics;
  \item A measure of the robot arm skeletal shape from base frame to end-effector;
  \item And a path-plan navigation algorithm.
\end{itemize}

\section{Proposed Computational Design Procedure}
\begin{figure}[htbp]
	\centerline{\includegraphics[width=60mm]{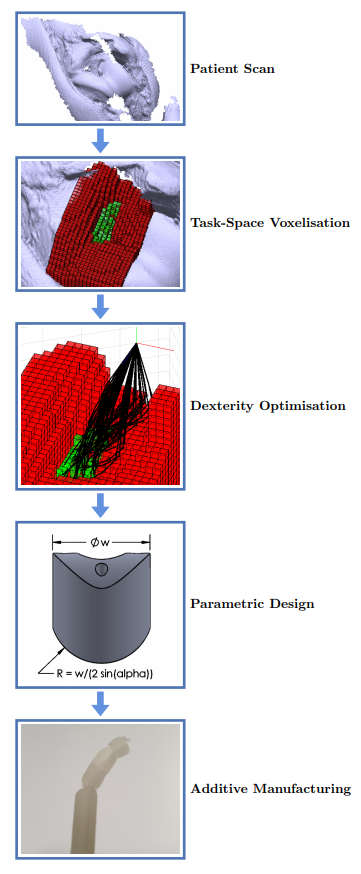}}
	\caption{Flowchart of the Design Procedure with the Patient scan at the top then Task-space voxelisation, the optimisation phase, parametric design and finally the 3D print model at the end.}
	\label{flowchart}
\end{figure}

The proposed procedure follows the patient-specific paradigm as visible in Fig. \ref{flowchart}. The processes of that procedure goes from a patient scan, to defining the task space as voxels, to optimizing the robot model for dexterity and finally to manufacturing a parametric design with its optimal parameters with 3D printing. The outcome is a dexterous patient-specific disposable tool for surgery.

\subsection{Task Space Data Voxelisation}

The first process is to turn 3D patient data into task space obstacles and regions of interest. This begins with collecting 3D data points of the task space or robot environment. For example in surgery, this could be an MRI, CT or ultrasound scan of the patient's anatomy before the operation. It is assumed that a scan like that is a fairly good representative of the task space in the actual operation. The next phase is to discretize the 3D data points into voxels and classify regions as obstacles or as regions of interest. This can be done through a neural network that can do 3D semantic segmentation of the scanned anatomy or with inspection by a human in a computer program. The result of this discritization is evident in Fig. \ref{flowchart} where the knee scan\footnote{Patient specific data acquisition was approved by the Australian National Health and Medical Research Council (NHMRC) - Registered Committee Number EC00171 under Approval Number 1400000856. We thank Andres Marmol for providing the knee scan dataset used in this paper.} was discretized with respect to the port frame for robot insertion which is assumed to be defined in this process as well. 

The last process is to dilate the obstacles and to fill in voids behind the region of interest as well as creating the voxel data file for the next process. 
Filling in voids as obstacles ensures that configurations cannot pass through the region of interest. Dilating is based on the thickness parameter for the robot where like in \cite{Bergeles2015}, this is to assume centre-line representations of the robot in the next procedure. Finally, the voxel information needs to be summarized in a data file where all the voxel coordinates, bounds, class labels and future dexterity measures shall be kept.

\subsection{Fitness Function: Sampling Configurations}

\begin{algorithm}
\caption{Fitness Function Dexterity in Region of Interest}\label{fitnessfunction}
\begin{algorithmic}[1]
\Function{FitnessFunction(parameters,Voxels)}{}
\State $\textit{JointLimits,n} \gets \text{RobotModel(parameters)}$
\For{$i = 1:n$}
\State $\textit{Q} \gets q_i = random(qi_{min},qi_{max},SampleSize)$
\EndFor
\State $\textit{Base Frame} \gets \text{Voxels}$
\For{$i = 1:SampleSize$}
\State $\textit{q} \gets Q(i,n)$
\State $T_{end}, X = \textit{ForwardKinematics(q,p,Baseframe)}$
\If {$\textit{Voxel}(T_{end}) \in V_{roi} \And \textit{Voxel}(X) \notin V_{obs}$}
\If {$\textit{ReachablePathPlan}(q_0,q_i) == \textit{True}$}
\State $P_z \gets R_{end} \gets T_{end}$.
\State $\theta,h = NearestPatch(P_z)$
\State $\textit{UpdateServiceSphere}(Voxel(T_{end}),(\theta,h))$
\EndIf
\EndIf
\EndFor
\State $N_{Vroi},N_{\theta},N_{h} \gets Voxels$
\For{$i = 1:N_{Vroi}$}
\State $surface patches \gets ServiceSphere(Voxel(i))$
\State $ No_{total} \gets N_O(i) = sum(surface patches)$
\EndFor
\State $\textit{Dexterity} = No_{total} / (N_{Vroi}N_{\theta}N_{h})$
\State $\text{Return} = \textit{-Dexterity}$
\EndFunction
\end{algorithmic}
\end{algorithm}

The fitness function defines the performance metric of the robot that shall be optimised in an optimization algorithm. The fitness function in this study is defined in algorithm \ref{fitnessfunction}. The idea is to evaluate different parameterised designs $p$ that consists of $m$ parameters in a design space $D$ and measure its fitness $F$ in the task space voxels $V$.
\begin{equation}\label{f}
F = f(p,V);   p \in D \subset \mathbb{R}^{m}
\end{equation}
To achieve optimal dexterity, we propose to use a Monte Carlo method to calculate the workspace to statistically analyze the dexterity distribution in the region of interest as the measure of the performance of each design. This method involves randomly sampling the joint space to find configurations that are representative of the configuration space $Q$. For a sample size of $N$, we get a matrix of $N \times n$ consisting of vectors $q_1$ to $q_N$ which are kinematic inputs for $N$ random configurations from the configuration space where $n$ is the number of DOFs for the robot model. For each $q_i$ from $q_1$ to $q_N$, we calculate the forward kinematics for the design which outputs the homogeneous transform from robot base frame to end-effector frame. 
\begin{equation}\label{forwardkinematics}
T_{end}^{base} = K(q_i,p);   q_i \in Q \subset \mathbb{R}^{n}
\end{equation}
In addition to this, we need to check if the configuration $q_i$ and the motion path-plan $\Pi$ to it is collision-free. We can do this by sampling points along the robot configuration from the base to the end-effector to check that the robot 3D shape $X$ is in a collision-free pose. These points represent the robot arm as a centre-line skeleton and can be integrated with the forward kinematics calculation. The sample size for shape should ensure that the sampled points are not further than the size of a voxel. 
For the path-plan, that would be a series of points for the robot shapes from configuration $q_0$ to $q_i$ under an obstacle avoidance path-planning algorithm. For some tubular snake-like robots (like the one used in the implementation section), the path-plan varies little to the shape of the configuration so the assumption that $\Pi_{q_0 \to q_i} \approx X_{q_i}$ is valid. However, the general case is to check that $X_{q_i}$ is collision-free before doing a path-plan check. 
To do our measurements for dexterity, we can use the above to form a condition to filter out configurations that fail to reach and path plan to the region of interest in a collision-free manner:
\begin{equation}\label{condition}
Voxel(T_{end}^{base}) \in V_{roi}   \And   Voxel(X) \notin V_{obs}
\end{equation}
where $Voxel()$ is a function that maps the Cartesian point to its corresponding voxel. If the end-effector reaches the target and the shape voxels are not obstacles, then we must ensure that the path-plan is possible with another condition:
\begin{equation}\label{Path plan condition}
ReachablePathPlan(q_0,q_i) == True
\end{equation}
where $ReachablePathPlan$ is a function that outputs a boolean if a successful motion exists in navigating from $q_0$ to $q_i$ without collision. 
If both of these conditions are true, then we can proceed with the dexterity measure; otherwise, we can ignore that configuration and move on to the next one.

\subsection{Fitness Function: Dexterity Index}

The measure for orientable dexterity in this study is based on the index developed in \cite{Liao2017}. 
To measure orientations for each specific spatial position (that is a voxel), each voxel contains a unit sphere called the Service Sphere. All of the possible areas on the sphere that can be oriented by the end-effector of the robot are referred to as Service Regions. This gives a dexterity measure for a voxel:
\begin{equation}\label{Dexterity1}
    Dex(v) = \frac{A_R(v)}{A_S} = \frac{N_O(v)}{N_{\theta}N_h} \in [0,1]
\end{equation}
Where $A_R(v)$ is the area of the Service Region in voxel $v$ and $A_S$ is the total surface area of the unit sphere. To measure the area of Service Regions on the unit sphere, we use a discretization method. The surface of the sphere is discretized into $N_{\theta}$ by $N_h$ equally sized patches along longitude meridians and latitude lines with the angle between adjacent meridians being $\delta \theta$ and the height interval of latitude lines being $\delta h$. Letting $N_0(v)$ represent the number of patches for the voxel that had been covered by the tooltip, we can calculate the dexterity $Dex(v)$ as the ratio of the surface patches reached by the tip $N_0(v)$ over the total surface patches of the Service Sphere $N_{\theta}N_h$. 
This ratio is bounded between 0 and 1 where 0 means that the tip could not reach any patch on the Service Sphere for that voxel and 1 means that the tip could reach all patches as the robot can touch the voxel from any orientation. 
So when the condition in Eq. (\ref{condition}) is true, we take the homogeneous transform $T_{end}^{base}$, extract the rotation matrix $R_{end}^{base}$, get the surface position from the z-axis unit vector $P_z$ to find the corresponding latitude and longitude coordinate $(\theta,h)$ in the unit sphere. 
We then update the Service Region for voxel $v$ and store the dexterity measure until all $N$ configurations have gone through this procedure. Once that is complete, all that is left is to find the total dexterity for the voxels in the region of interest. This would be a sum of all dexterity measures for each voxel $v$ in $V_{roi}$:
\begin{equation}\label{Dexterity2}
    Dex(V_{roi}) = \int_{v\in V_{roi}} \frac{Dex(v)dv}{V_{roi}}  \in [0,1]
\end{equation}
As a discrete measure this is a sum of all Service Region patches $N_O$ in $V_{roi}$ over the total area of the Service Sphere $N_{\theta}N_h$ divided by the total number of voxels in the region of interest $N_{Vroi}$:
\begin{equation}\label{Dexterity3}
    Dex(V_{roi}) = \frac{\sum_{i=1}^{N_{Vroi}}  N_O^{i}}  {N_{Vroi}N_{\theta}N_h} \in [0,1]
\end{equation}
This total dexterity measure is the main objective to maximize. Since most standard algorithms minimize functions, the measure is treated as a negative for the output of the fitness function:
\begin{equation}\label{Output}
F = -Dex(V_{roi}) \in [-1,0]
\end{equation}
It is possible to add multiple objectives such as a penalty for the complexity of design parameters with a weighted sum of importance for the output of $F$ in Eq. (\ref{f}). For this study, only the dexterity objective shall be used for the implementation.

\section{Implementation of the Computational Design Algorithm for Designing the Manipulator}

This computational design algorithm was implemented for the application of designing a surgical manipulator. A snake-like manipulator is desired for the development of a teleoperated macro-micro robot for knee arthroscopy. This manipulator needs to pick and cut tissue in a certain region of cartilage. The Snake-like manipulator will be a micro robot attached to the end of the Raven II robot which is the macro robot. The Snake-like robot shall be manufactured cheaply through 3D printing so that it can be patient specific and disposable after the operation.

\subsection{Snake-like Robot Model for the Experiment: A Variable Neutral-Line Mechanism}

The variable neutral-line mechanism described in \cite{Kim2014} proposed design improvements over traditional snake-like manipulators that use a flexible backbone in tendon driven robots. Their design proposed a rolling joint set-up such that the robot mechanism had an adjustable stiffness. Their design was also hollow which is significantly important for attaching tools to the end of the robot such as a gripper or a camera. Therefore, for its many advantages, this design was used as the basis for developing the snake-like robot for this study.

The forward kinematics for this robot is described in \cite{Lee2014} as a series of the pan and tilt rolling joint transforms. For the parametric design for this mechanism, the fewest amount of variables to define the model were $n, d, w$ and $\alpha$ where $n$ defines the number of disk joints, $d$ is the joint height, $w$ is the width of the robot while $\alpha$ is half the rolling joint angle of curvature. These four parameters define the set of variables for a single segment or module. For multiple segments, The variable $w$ should be constant as the model would not be practical having a variable thickness for each segment. 
In this study, we will reduce our design space by defining $w = 4mm$ based on a suitable size for the robot to house an instrument and fit the trocar. This gives us $3 \times s$ parameters to solve. There are also some other parameters defined for the transition between multiple segments. In \cite{Lee2014}, those parameters are $\zeta$ and $\sigma$ which define the transition rotation and translation. To simplify this, we assume that the transition is integrated with previous joint such that $\sigma = d_{proximal}$ and that $\zeta = -\pi/s$ to ensure the tendons for each segment are evenly positioned. Each segment of the robot has two DOF which are pan and tilt joints. This gives a configuration space of five DOF for the single segment and seven DOF for a double segment.

\subsection{Differential Evolution Optimisation Set-up}

Since the fitness function is numerical, we need a good derivative free generic optimization algorithm to optimize for it. For this, differential evolution was chosen as the algorithm to optimize the function for its advantages over conventional genetic evolution algorithms \cite{Storn1997}. It is a parallel direct search method which has a scheme that helps the method converge faster to the optimal than adaptive simulated annealing and annealed Nelder-Mead approaches. 
For this study, the first strategy of differential evolution was used. This scheme generates trial parameter vectors by adding a weighted difference vector between two population members to a third member as shown in Eq. (\ref{DE1}). 
For $x_{r_i,G},i = 0,1,2,...,NP-1$, a trial vector $v$ is generated according to:
\begin{equation}\label{DE1}
v = x_{r_1,G} + F \cdot (x_{r_2,G}-x_{r_3,G}).
\end{equation}
For this study, we used a differential weight of $F=0.8$ and crossover value $CR = 0.7$ and the populations $NP$ being $10 \times p$ which was 30 for the single segment and 60 for the double segment. The evolution stop time was set for 24 hours with 10 test repeats.

\section{Experimental Results}

\begin{figure}[htbp]
	\centerline{\includegraphics[height=120mm]{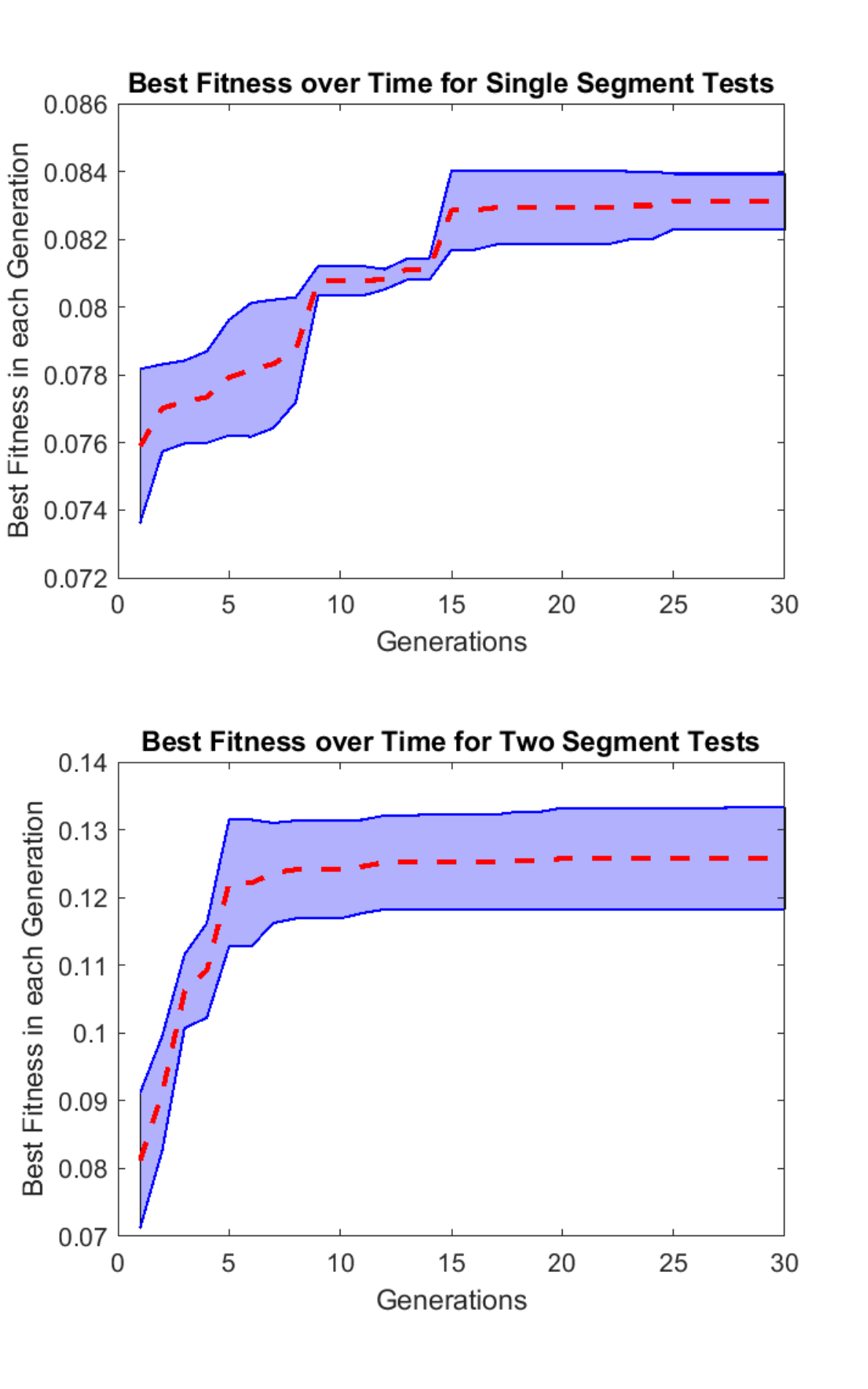}}
	\caption{The convergence of the differential evolution algorithm for the single and double segment optimization where the red line shows the mean fitness value for the 10 tests and the blue region marks the first standard deviation from the mean}
	\label{converge}
\end{figure}

\begin{figure*}[htbp]
	\centerline{\includegraphics[width=180mm]{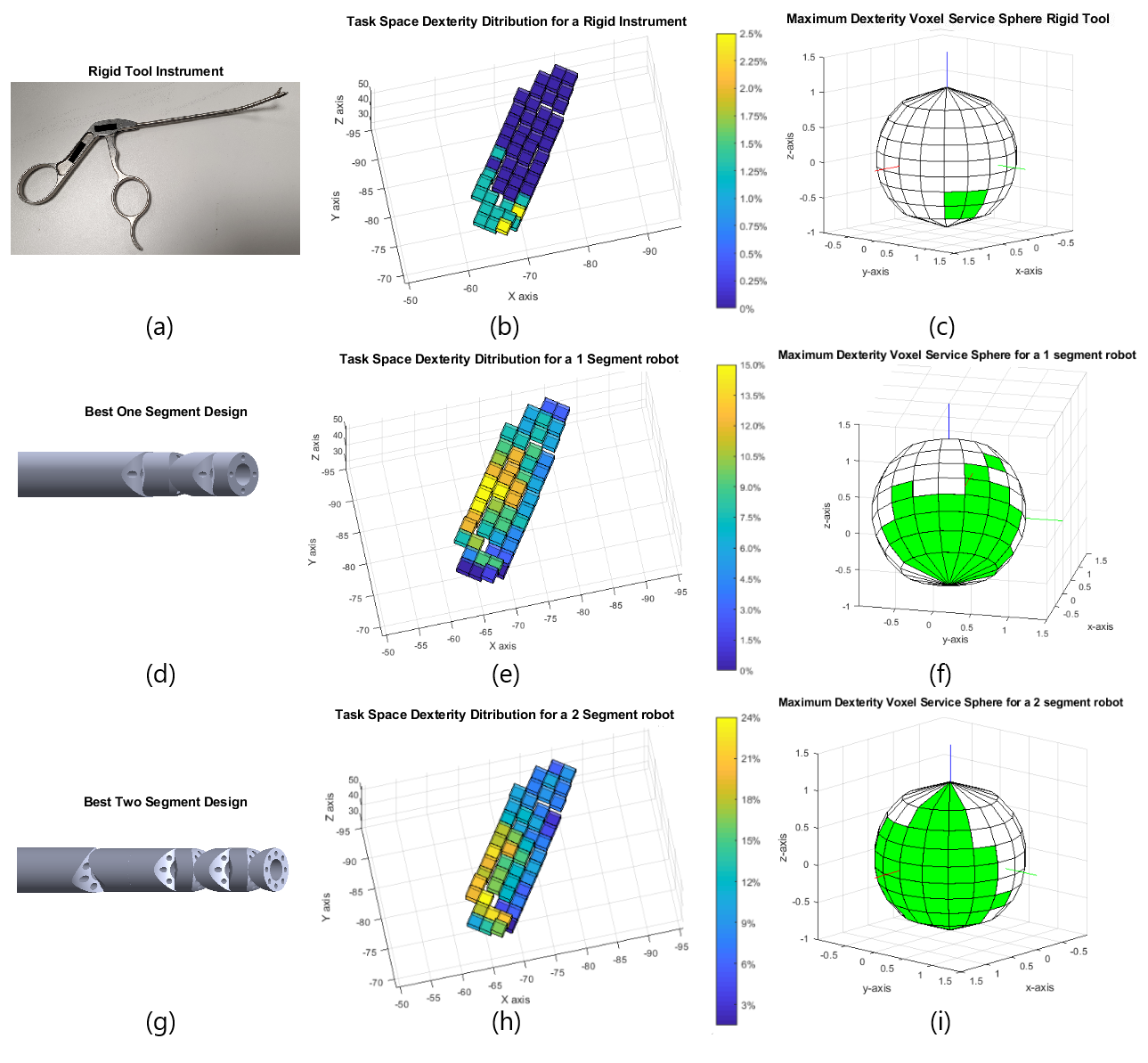}}
	\caption{Experimental results. (a) The rigid tool, (b) its dexterity distribution in the region of interest, and (c) its maximum dexterity service sphere. (d) The best one-segment design, (e) its dexterity distribution in the region of interest, and (f) its maximum dexterity service sphere. (g) The best two-segment design, (h) its dexterity distribution in the region of interest, and (i) its maximum dexterity service sphere.}
	\label{results}
\end{figure*}

The optimization was done for a single- and a double-segment design, respectively. 
The corresponding mean and maximum voxel dexterity values were compared with the fitness function results for a rigid instrument. The average evaluation time of the fitness functions for 1 segment was about 1 minute and 23 seconds and for 2 segments it was about 2 minutes and 21 seconds. The sample size was chosen as 1,000,000, the voxels were $2mm$ cubes and the surface patches were $N_{\theta} = 18, N_h = 9$. The robot also had a $5mm$ long tool attached to the end-effector. The optimization was done as a job for a High Performance Computer\footnote{Computational resources and services used in this work were provided by the HPC and Research Support Group, Queensland University of Technology, Brisbane, Australia} with a timer of 24 hours. The convergence of the optimization algorithms are shown in Fig. \ref{converge}. The designs and their dexterity measures are summarized in Tab. 1 and Fig. \ref{results}.

\begin{table}[h!]
\label{table:1}
\caption{Table of Experimental Results}
\centering
{\makegapedcells
\begin{tabular}{ |p{20mm}||p{15mm}|p{15mm}|p{15mm}|  } 
 \hline
 \multicolumn{4}{|c|}{Experimental Results} \\
 \hline
 Model & Parameters & Mean Dexterity & Maximum Dexterity \\
 \hline
 Rigid Tool   &  NA  &   0.0055  &   0.0247 \\
 \hline
 Single Segment $\alpha, d, n$ & 1.24, 1.62, 3  & 0.0835   & 0.1358\\
 \hline
 Double Segment $\alpha_1, d_1, n_1,$ $\alpha_2, d_2, n_2$ & 1.34, 6, 1, 1.18, 0.41, 3 &  0.1351  &  0.2272\\
 \hline
\end{tabular}}
\vskip 1mm
\end{table}

\section{Discussion}
In general, the pattern from the results in Fig. \ref{results} shows that the more DOF the robot has the better the dexterity. The service sphere and distribution for the rigid instrument in Fig. \ref{results} shows that its orientable dexterity is much less than the optimal snake-like robots. It is intuitive to expect that a rigid tool would have a very limited range of orientability about a small voxel from a single port. By its distribution, it is also evident that it was designed to manipulate tissue from the furthest end of the region of interest from the port. The single segment, on the other hand, shows a much better service sphere and a more evenly distributed dexterity in the region of interest. The evolution algorithm chose to have $n = 3$ and a large $\alpha$ angle. This choice leads to having a snake-robot with a small wrist-like joint for fine manipulation as the Raven II provides the workspace coverage. Lastly, the double segment has the greatest service sphere and a higher valued distribution in the region of interest. The evolution interestingly chose to have a single proximal joint and a few smaller distal joints. Note that this means that the design lacks a DOF as the proximal section is a single pan joint giving a total of 6 DOF in the design. This verifies that 6 DOF is sufficient for this dexterous task. Furthermore, this design resembles the structure of a human arm because the Raven II acts as a shoulder joint, the proximal joint behaves like an elbow for a wider workspace and the distal joints behave like a wrist for fine manipulation. Clearly, the algorithm chose these features as they desired the best task-space coverage and orientability for each voxel in the region of interest. 
To further verify that the two-segment design is significantly better than the one-segment design in terms of performance, a Mann-Whitney U test was used to compare the designs and their dexterity results for the 10 tests in Fig. \ref{converge}. The Z-score from that was 4.3487 indicating a strong $99.99\%$ level of confidence that the two segment design performs better than a single segment. 

Implementation of the computational design algorithm has involved tuning many settings that may not be intuitive to justify. It is unclear how accurate the simulation of the task space should be to produce a sufficient design. Changing settings like increasing the sample size, decreasing the patch size and the voxel size all make the computation exponentially more expensive. Tuning these values can make this algorithm difficult to adapt to the practical use of developing patient-specific robots. 
Further work can be done to improve the algorithm computationally with parallel loops. The current challenge in doing this is with updating the same voxel service spheres simultaneously in Algorithm \ref{fitnessfunction}. 
Other improvements can be to integrate a path plan checker using sampled configurations to define a reachable map. Another improvement can be to combine other objectives and global optimisation for the port location and the other tools in the operation. Finally, future work in using the design in a teleoperation task is needed to observe how effective the dexterity performance is for enhancing the surgeon's abilities in the application.

\section{Conclusion}
In this study, tendon-driven pseudo-continuum robots were designed under the patient-specific paradigm for knee arthroscopy. A novel algorithm was proposed to optimize the design for optimal oriental dexterity within a confined task space. It was implemented for a variable neutral-line mechanism for a tendon-driven robot. The results show that the optimal snake-like robot designs had significantly greater dexterity than a conventional rigid tool used for the application. Overall, the results show that this algorithm is feasible for the design of surgical manipulators under anatomical constraints but further work is required to improve the algorithm and to find ways of adjusting the settings for different surgical applications.

%\textcolor{red}{You can use a separate bib file including all the entries of each reference, so that the format follows IEEE requriements.
%Entries of each reference can be copied from google scholar. - Leo}

\bibliographystyle{IEEEtran} 
\bibliography{root}

\end{document}